\RequirePackage{fix-cm}
\documentclass[smallextended]{svjour3}       
\smartqed  
\usepackage{graphicx}
\usepackage[numbers]{natbib}
\begin{document}

\title{An Attribute Oriented Induction based Methodology for Data Driven Predictive Maintenance
}


\author{Javier Fernandez-Anakabe*         \and
        Ekhi Zugasti Uriguen \and
        Urko Zurutuza Ortega 
}


\institute{Javier Fernandez-Anakabe \at
              Mondragon Unibertsitatea - Goi Eskola Politeknikoa, Loramendi 4, 20500 Arrasate - Mondragon (Gipuzkoa) \\
              \email{jfernandeza@mondragon.edu}      \and
           Ekhi Zugasti Uriguen \at
              Mondragon Unibertsitatea - Goi Eskola Politeknikoa, Loramendi 4, 20500 Arrasate - Mondragon (Gipuzkoa) \\
              \email{ezugasti@mondragon.edu}
            \and
            Urko Zurutuza Ortega \at
            Mondragon Unibertsitatea - Goi Eskola Politeknikoa, Loramendi 4, 20500 Arrasate - Mondragon (Gipuzkoa) \\
            \email{uzurutuza@mondragon.edu}
}

\date{Received: date / Accepted: date}

\maketitle

\begin{abstract}
Attribute Oriented Induction (AOI) is a data mining algorithm used for extracting knowledge of relational data, taking into account expert knowledge. It is a clustering algorithm that works by transforming the values of the attributes and converting an instance into others that are more generic or ambiguous. In this way, it seeks similarities between elements to generate data groupings. AOI was initially conceived as an algorithm for knowledge discovery in databases, but over the years it has been applied to other areas such as spatial patterns, intrusion detection or strategy making. In this paper, AOI has been extended to the field of Predictive Maintenance. The objective is to demonstrate that combining expert knowledge and data collected from the machine can provide good results in the Predictive Maintenance of industrial assets. To this end we adapted the algorithm and used an LSTM approach to perform both the Anomaly Detection (AD) and the Remaining Useful Life (RUL). The results obtained confirm the validity of the proposal, as the methodology was able to detect anomalies, and calculate the RUL until breakage with considerable degree of accuracy.
\end{abstract}

\section{Introduction}
\label{intro}
Maintenance has been a constant issue in industry as industrial machines tend to suffer wear over time, either due to natural operational processes, or damage caused. Taking this into account, companies aim to save time and money, constantly seeking reduction in the cost of their products and services. Key to this is improving productivity.

Over the years, strategies to manage maintenance tasks have been evolving. This evolution has caused a shift in thinking, from maintenance being a necessary evil, to becoming a critical driver of competitivity. More importantly, maintenance has evolved from being an expense to becoming an investment.

Science and technology have evolved at great speed enabling the development of elements that allow data capture and storage in vast quantities. The industrial sector has taken advantage of these technological advances, with companies increasingly implementing predictive maintenance methodologies to improve productivity. 

In the last decade, data monitoring and collecting have become mainstream in the majority of companies of the industrial sector (generally based on the application of Machine Learning tools) to improve productivity.

There are several types of maintenance, but R. Keith \cite{aitpm} highlights three main types:

(i) The purpose of \textit{\textbf{Corrective Maintenance} is to manage and schedule repairs after a failure has occurred.} Until recent years, this has been one of the most commonly used maintenance strategies. Corrective Maintenance is a methodology that is based on executing a correction or maintenance action when a breakage occurs in the asset that is being monitored \cite{pmmftles, pmdtmomacs}. In this way, maintenance efforts are minimised, allowing the asset to function until an error occurs or cannot provide more service. The disadvantage of this methodology is that it does not take into consideration assets suffering a degradation which, even if no error occurs, can prevent it from working efficiently. Therefore, no maintenance is scheduled, thus costs soar once breakage happens.

The amount of time that a machine is not working properly or is stopped due to a breakdown incurs loss of time and money for the company. Thus, over the years more sophisticated methodologies have been developed.

(ii) \textit{\textbf{Preventive Maintenance} tries to schedule maintenance tasks based on time periods.} This strategy is based on establishing maintenance periods a priori, to revise the state of the asset or machine \cite{pmdtmomacs, cbmomt-ar, csmpcbm-r, aootbacbmia}. These periods are generally established by domain experts, who have the most specific knowledge of the product. More importantly, maintenance has gone from being an expense to being an investment. One drawback of this methodology however, is the out of period wear or breakdown are not taken into account. Furthermore, scheduled maintenance can also lead to parts being replaced earlier than necessary.

(iii) \textit{\textbf{Predictive Maintenance} is focused on preventing unscheduled downtimes and premature damage to equipment.} It is based on capturing data from the asset to be monitored, and extracting relevant information. By using a data history that has been recording the behaviour of the asset, it is possible to predict when the next anomaly could occur, and thus, prepare a preventative maintenance action. In this way, estimates of maintenance times and actions can be obtained more efficiently and accurately than by employing Corrective and Preventive Maintenance strategies.

Nevertheless, there is another type of maintenance called Proactive Maintenance, which aims to analyse the state of the machine and perform actions to ensure its useful life. The difference with Predictive Maintenance is that it not only tries to analyse when the next failure will occur, but also suggests how to manage that situation to obtain optimal machine health \cite{foanpmfspmiois, omapmds, tMANTISb}.

\begin{figure}[h!]
  \centering
  \includegraphics[width=0.8\textwidth]{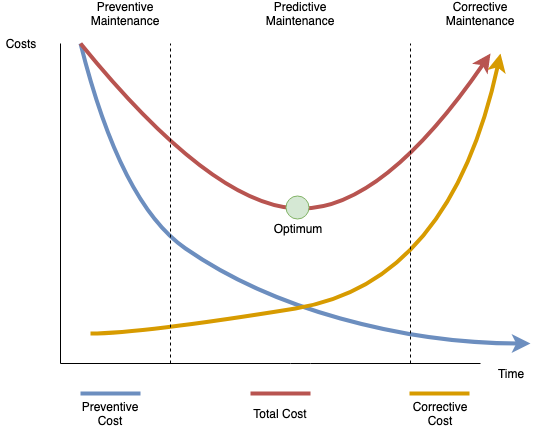}
  \caption{Cost relation between the different maintenance strategies \cite{rpapm}.}
  \label{Maintenance-costs}
\end{figure}

Figure \ref{Maintenance-costs} shows the cost relationship between these three maintenance methodologies. While over time the repair cost increases due to the level of degradation suffered by the machine, the cost of prevention decreases. Thus, the later the machine is replaced, the greater the profitability. Moreover too early a substitution can result in significant losses. It is important to determine the optimum point at which maintenance must be carried out, and this can be achieved through Predictive Maintenance.

\begin{figure}[h!]
  \centering
  \includegraphics[width=0.8\textwidth]{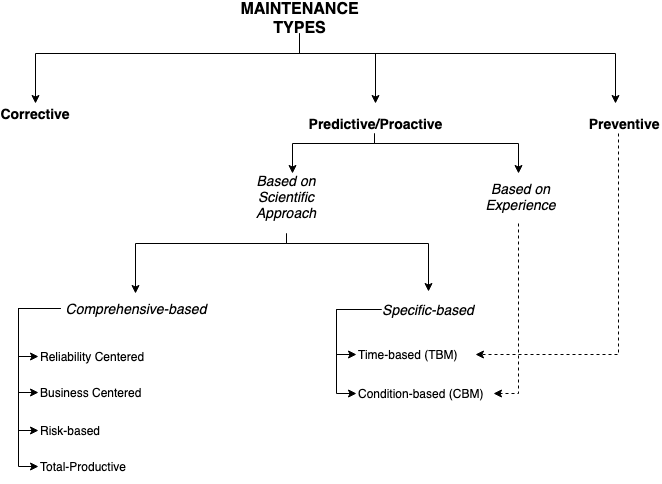}
  \caption{Distribution of the different maintenance types.}
  \label{Maintenance-types}
\end{figure}

The different maintenance strategies are reflected in Figure \ref{Maintenance-types}.

Predictive maintenance is made up of three elements: (i) Anomaly Detection (AD), (ii) Root Cause Analysis (RCA), and (iii) Remaining Useful Life (RUL). Inside these elements, there are different data analysis techniques to process the information collected by sensors, and turn it into knowledge.

This paper has focused on the use of an algorithm called Attribute Oriented Induction (AOI) to perform the calculation of the different components. The novelty resides in the fact that this algorithm has been used for the first time in a Predictive Maintenance scenario.

The output of the AOI algorithm is a set of clusters with a weight. Utilising those weights a quantification function was defined for each simulation that represents to the wear process of the monitored machine. Utilising this quantification function, AD and RUL were calculated, based on different thresholds and auxiliary methods, such as EWMA and LSTMs.

The paper is organised into the following sections: 1.Introduction, where the concept and evolution of Maintenance are set out. 2.Background, in which the main concepts of Predictive Maintenance and the AOI algorithm are explained. 3.Dataset description and preprocessing actions. 4.Methodology and Data Preparation for Experiments. 5.Results. 6.Conclusions and Final Remarks.

\section{Background}

In this section Anomaly Detection and Remaining Useful Life are explained, and AOI algorithm is described.

\subsection{Anomaly Detection}

Anomaly Detection (AD) is a field which started in the XIXth century, and over time, techniques have been improving. It can be defined as finding unusual behavioural states in the monitored asset that can be considered as anomalies. First of all, however, the meaning of the term \textit{anomaly} must be explained.

Anomalies are patterns in data that do not conform to a well defined notion of normal behaviour. In this sense, if the behaviour is not normal, this state may be considered a failure. So this fact makes a need of differencing an anomaly from a change state. A change state is a moment in the behaviour in which an uncommon procedure is detected, but this does not imply an error or incorrect behaviour. It is only a deviation from normal behaviour that can carry out the detection of a fault or anomaly.

Anomalies might occur in the data for a variety of reasons. One example of this is malicious activity, such as credit card fraud, cyber-intrusion, terrorist activity or breakdown of a system. Some experts define the anomaly detection as \textit{novelty detection}, which aims at detecting previously unobserved (emergent, novel) patterns in the data (Chandola et. al. \cite{chandola}).

It is clear that to build a solid anomaly detection system it is necessary to have a solid and reliable database. The more data you have, the better the chances of developing an effective anomaly detection solution. To this end, some approaches based on Machine Learning and Statistical domains have emerged.

Several methods based on Machine Learning are distinguished to detect anomalies. (i) Classification: where the data is labelled as normal or anomalous, and the idea is to try to train a model that can distinguish its label. (ii) Clustering: using techniques such as the K-Nearest Neighbours (K-NN) algorithm, groupings of data are generated, and those that do not fit the parameters are considered as anomalies. It can also happen that the data related to anomalies form clusters, but these are usually less populated than those that generate the normal elements.

Applications of Anomaly Detection: (i) Intrusion Detection: semi-supervised and unsupervised algorithms are the most used in this field \cite{aabdmfi, adondulpd, mmids}. They are usually datasets composed of large volumes of data and frequently there is usually a large false alarm rate. (ii) Fraud Detection: A profile is maintained for each user and controlled for any unusual movement \cite{ccfdwann}. (iii) Medical Anomaly Detection: a high level of accuracy is required as mistakes can impact lives and well being. Such cases are usually semi-supervised problems in which the labelled data belongs to healthy patients \cite{atmfumte}.

This study focuses on the detection of anomalies in the industrial sector \cite{diialowtd, bddadiiaaaacs}. In this field, monitored machines suffer damage over time, either by natural wear and tear, defects or external incidents. To prevent breakdowns, it is therefore very useful to detect when a machine is malfunctioning or is close to failure. This means that the data must be analysed in real time so that the detection of anomalies is as immediate as possible.

Chandola et. al. \cite{chandola} observed that the detection of anomalies can be sequential. Thus it makes sense considering that most of the wear data of a machine is related to time. The prediction of when an anomaly will occur, will be made with time series analysis.

To clarify the terminology in this study, two distinct processes will be distinguished: (i) change detection, in which the objective is to detect deviations in the process; and (ii) anomaly detection, whose objective is to find when a state can be considered as failure or breakdown.

\subsection{Remaining Useful Life}

One of the goals of Predictive and Proactive Maintenance is to try to estimate how much time an asset or a service is able to continue functioning until a failure or anomaly occurs.

Modern day companies cannot afford the loss of time and money due to an unexpected (and sometimes even expected) machine- or service-failure. For this reason strategies and schedulings to prevent and predict stoppages have become necessary.

The process of foreseeing the resting time until a failure occurrence is named Remaining Useful Life (RUL).

Si, X.-S. \cite{rulearotsdda} defined the concept as: \textit{the useful life left in an asset at a particular time of operation}. He noted that the estimated time until the normal behaviour of an asset finishes is typically unknown. Information obtained from condition and health monitoring is relevant to help determine the \textit{RUL}, and thus the concept is strongly related to Predictive Maintenance. On the other hand, Okoh, C. \cite{oorulptitles} proposed the following definition to describe RUL: \textit{Remaining Useful Life (RUL) is the time remaining for a component to perform its functional capabilities before failure.} Even though many authors do not include the term \textit{failure} in the description, they agree in defining the concept as the end of the correct-functional capacity of an asset.

There are multiple studies in the literature referring to the calculation of RUL. Ahmadzadeh, F. and Lundberg, J. \cite{ruler} made a review of the different techniques to estimate RUL, and distinguished four different methodologies: (i) \textit{physics based models} construct technically comprehensive theoretical models to describe the physics of the system; (ii) \textit{experimentally based models}, use probabilistic or stochastic models of the degradation phenomenon or the life cycle of components, by taking into account the data and the knowledge accumulated by experience; (iii) \textit{data-driven models} are based on processing the collected data, without needing special product knowledge to be specified; and (iv) \textit{combination/hybrid prognostic methodologies}, which are based on combining more than one of the previously mentioned models.

The most discussed and recommended Machine Learning algorithms to estimate RUL observed in the literature are \textit{ARIMA} models \cite{copaferulob, eafomhcuam}, Artificial Neural Networks (\textit{ANN}) \cite{aannafrulpoestcm, ruler, oorulptitles, annafrulpubfash, prulormbann, tuomannnfatrulob, rulpormuhdnn}, \textit{Support Vector Machines} (SVM) \cite{copaferulob, ruler, oorulptitles, mpdaarulpuphmas, mpbohseusvm}, Bayesian approaches \cite{copaferulob, oorulptitles, ruler} and Hidden and Semi-hidden Markov Models (HMM, or SHMM)\cite{ruler, oorulptitles}.

Ahmadzadeh, F. \cite{ruler} affirmed that the use of prognostics in calculation of RUL is crucial, and analysed the capacity of multiple algorithms to address this, utilising recent RUL estimation applications from the literature. The ANN, SVM and Bayesian Approaches are some examples from this comparative study into the field of data-driven approach, and \textit{ARMA} family models if a Hybrid-model is considered, combined with any other Machine Learning algorithm such as Bayesian Networks or ANN.

\begin{figure}[h!]
  \centering
  \includegraphics[width=0.8\textwidth, height=6cm]{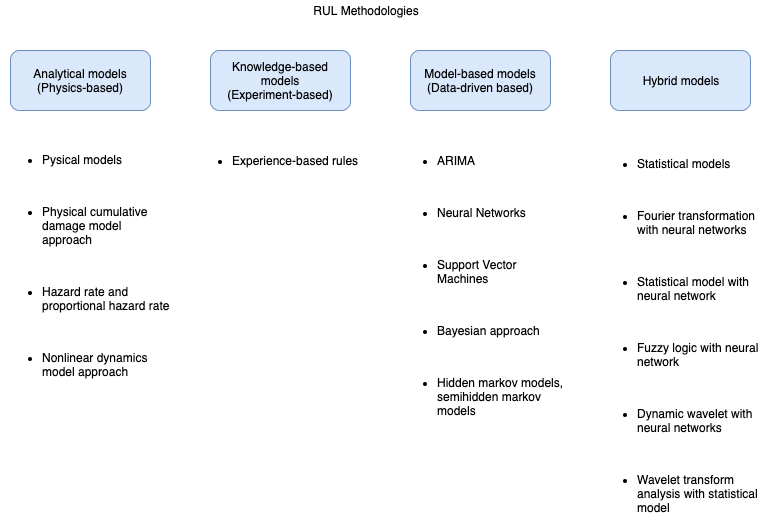}
  \caption{Classification of the different RUL methodologies according to their characteristics.}
  \label{RUL-met}
\end{figure}

The aforementioned studies can be summarised as in Figure \ref{RUL-met}.

\subsection{Attribute Oriented Induction}

Attribute Oriented Induction is a hierarchical clustering algorithm, and was first proposed by Jiawei Han, Yandong Cai, and Nick Cercone in 1992 as a method for knowledge discovery in databases \cite {knidaaoa, erbaoifdm}. It structures concepts in different hierarchies, called \textit{generalisation levels}. Each of the attributes that compose the data have their own structure of generalisations or \textit{hierarchy-tree}. When the values of an attribute are transformed into values of another level of generalisation it is called \textit{concept-tree ascension} \cite{erbaoifdm}.

\begin{figure}[h!]
  \centering
  \includegraphics[width=0.8\textwidth]{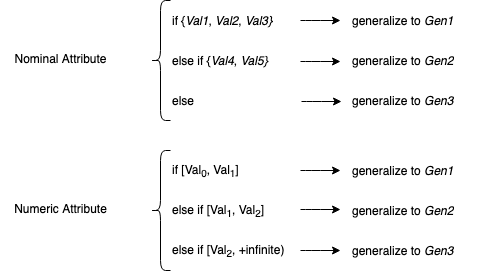}
  \caption{Example of the representation of generalisation-rules for nominal or numerical attributes.}
  \label{aoi-nominal-numeric}

\end{figure}

To ensure the correct functioning of the algorithm it is necessary to establish background knowledge in which the different generalisation levels of each attribute are specified, as shown in Figure \ref{aoi-nominal-numeric}. Background knowledge is a set of rules and terms that provide contextual information of the scenario.

\begin{figure}[h!]
  \centering
  \includegraphics[width=0.8\textwidth]{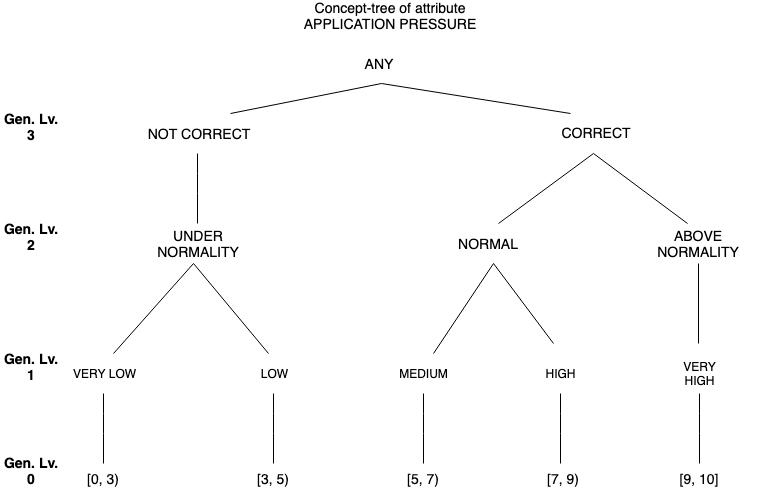}
  \caption{Example of a concept-tree or concept-hierarchy of a Predictive Maintenance attribute: Application Pressure.}
  \label{concept-tree-aoi}

\end{figure}

An example of the hierarchy-tree (also called \textit{concept-hierarchy}) \cite{knidaaoa} is shown in Figure \ref{concept-tree-aoi}. The background knowledge should ideally be defined by domain experts. Domain experts are the most experienced people on the area of study. This information should be useful to represent the health status of the monitored machine, as accurately as possible. The AOI algorithm needs to use background knowledge to represent generalisations, so that the more reliable the information, the more representative the results will be for experts.

AOI has been applied in many scenarios, such as spatial patterns, medical science, intrusion detection, strategy making, and financial prediction \cite{WU2011134, admafaowatasg, cidatsrca}. However, after reviewing the literature, no reference was found indicating that AOI has been used in scenarios such as Predictive Maintenance. The ability of this algorithm to combine data collected from the machine and information provided by experts in the field, can prone vital to assisting operators in making better maintenance decisions.

Although AOI algorithm was used in different contexts or applications, none of them were real-time processes, nor their purpose was the evaluation of AD or RUL. Thus, one of the objectives of this research project is to validate its usage to reduce the required computation force and extract knowledge from raw data.

\subsubsection{Main features of AOI and how it works}

Data selection and preprocessing is one of the most important steps to obtain benefits from AOI. The power of AOI resides on defining thresholds and generalisation hierarchies for the features of the data. This is why the application of the AOI must be managed in close collaboration with domain experts. The goal is to establish the most precise generalisation criteria, and thus obtain the most reliable representation of the data after the generalisation is performed. 

There are some parameters that must be set before starting the cluster generation process:

\begin{itemize}
\item \textit{\textbf{Minimum Cluster Size}: The minimum number of instances (aka. tuples) needed to form a cluster.}

\item \textit{\textbf{Attribute generalisation Threshold}: The maximum number of instances allowed in a generalisation level. If the number of different values of the generalised-level of an attribute is greater than the specified threshold, a generalisation must be applied to that attribute.} 

\item \textit{\textbf{Tuple generalisation Threshold}: The maximum number of distinct tuples allowed in a generalised relation. If the number of different tuples in a generalised relation is higher than the threshold, a generalisation must be applied to one of the attributes. Different criteria can be used to select the attribute to be generalised.}

\item \textit{\textbf{generalisation Hierarchies and Rules (for each attribute)}: Definition of different generalisation levels for each attribute with their respective elements; and the conditions to determine which value of the next generalisation level an element can be generalised to.}

\item \textit{\textbf{Attribute generalisation Weights}: Each attribute must have a specified weight according to their relevance in the clustering process. These weights must be specified by the domain expert to guarantee the coherence of the weighting proposal. This step is important to ensure the correct working of the AOI clustering process.}

\end{itemize}

Once the preprocessing-parameters are set, there are several steps or strategies that must be followed to work with the traditional AOI algorithm, which are listed below:

\begin{enumerate}

\item generalisation of the smallest decomposable attributes: The most heterogeneous attribute (with the highest number of different values) is the one that must be generalised to reduce the diversity of the dataset and help minimise the complexity, in accordance with the Least Commitment Principle \cite{friedmanleast}.

\item Attribute removal: If there is an attribute with more distinct values than the generalisation Threshold, and there is not a higher generalisation level for it, the attribute must be removed. 

\item Concept tree ascension: If there a higher level generalisation exists in the generalisation hierarchy for an attribute, all the values of the current generalisation level of that attribute must be replaced by the next level values. 

\item Vote propagation: When an attribute is generalised, some tuples become identical. In such cases the votes from the tuples on the previous generalised relation must be added to the new generalisation. At the beginning of the clustering process, the votes of each tuple are set to 1. 

\item Threshold control for each attribute: If there is an attribute with more distinct values than the Attribute generalisation Threshold, then the attribute must be generalised to the next level of the hierarchy.

\item Threshold control for generalised relations: If there is a generalised relation with a higher number of different tuples than the Tuple generalisation Threshold, further generalisation of an attribute must be performed. In this approach, the attribute that will be selected to be generalised is that with the highest number of different values. In case of more than one attribute with the same number of different values occurring, the one with the highest weight will be chosen.

\end{enumerate}

The AOI algorithm overcomes two major obstacles. On the one hand, it characterises the origin of an anomaly or error. This is the foundation of Root Cause Analysis, since it extracts knowledge and gives meaning to raw data. On the other hand, AOI is able to summarise vast amounts of data into a small number of groups.

The AOI algorithm follows a cyclic methodology. It first generalises the most changing attribute (the attribute with the highest number of different values). Next, it generalises the second most changing attribute. It continues the generalisation process until it meets a stopping criteria, like the absence of sufficient similar elements to form a cluster. Finally, it provides a collection of data groups or clusters. Moreover, every step, the data processing workload is reduced.

In the execution of the generalisation process, first of all the most heterogeneous attribute is detected. Then the attribute with the highest number of distinct values is generalised to the next level, following the hierarchy-tree criteria established previously. When an attribute is generalised, the system tries to find similar rows of data. If there are a number of similar rows higher or equal to the number of elements needed to form a cluster, those elements are removed from the table, and a cluster is generated. The execution continues until no more elements are in the table, or no more clusters can be generated from the remaining elements.

\subsubsection{Variants of AOI}

In this study, some variants have been added to the traditional AOI algorithm.

The first of these is the inclusion of a loop in the execution of the algorithm. One of the parameters that must be specified prior to the execution of AOI is the minimum number of instances that will generate a cluster. This process is quite arbitrary and depending on which value is specified, there may be instances that do not belong to any cluster at the end of the clustering execution. For this reason, an iterative process has been defined in which the value of the minimum cluster size decreases. That is, suppose we set a value for the minimum cluster size of 10. The algorithm runs with that value and at the end there are X elements that are not clustered. In that case, the AOI algorithm would again be executed with a minimum cluster size value of 9 with the remaining elements (or X in the case of $X < 9$). The process continues until the value is 2. In this way, the number of unclustered elements is minimised.

The second variant included is the addition of a weight to each cluster. This is to perform the calculation of AD and RUL. It is logical to think that a cluster that belongs to a generalised relation in which the levels of generalisation are low refers to more representative states than those whose levels of generalisation are greater. For example, suppose there is a cluster in which none of the attributes has been generalised. In this case, it is clear that the similarities at the time of generating the cluster have been based on raw data. Therefore, it is considered that this cluster refers to a habitual and specific state of the behaviour of the machine. If we have a cluster that has some attributes that have been generalised to a higher level, the representation of the state is no longer as accurate, in fact it is more abstract.

It is important to take into account the number of elements that form the clusters. It is not the same for a specific cluster to be made up of 2 or 200 instances. The greater the number of elements that composes it, the greater the weight, and this will be affected by the level of generalisations of the attributes. The process of assigning weights to clusters is explained in the section Methodology and Data Preparation for Experiments.

Moreover, some other variations present in the literature must be mentioned. Those papers are related to the implementation and usage of the Attribute Oriented Induction High Level Emerging Pattern (AOI-HEP). Spits Warnars \cite{aoiohlep}argued that the AOI algorithm does not discover new emerging patterns. Therefore implementations of AOI-HEP have been carried out to analyse the capacity of the algorithm to manage emerging patterns on data based on the rules obtained after the processing of the AOI \cite{mfpwaoihlep, uaoihleptmfp, marwaoiglepdmt}.

In the present paper, LSTM models have been selected due to their capacity to manage different sources of data to make predictions. Other models such as ARMA family models can be utilised too, but the decision has been taken due to the LSTM can take into account more parameters of the input data.

\section{Dataset}

The dataset utilised to carry out the experiments was provided by the Prognostics CoE at NASA Ames \cite{tweod}. This dataset is made up of multiple simulations of the wear in a turbine engine. There are two groups of data: (i) training data with 100 different simulations, starting at a random point in the wear process, and ending after the error has occurred, and (ii) test data with 100 different simulations, starting at a random point in the wear process and ending before the error occurs. It should be noted that each start moment of each simulation can be different, and it is not specified in any of them. In the case of the test dataset, the number of work cycles before the failure occurs is also not registered. The type of failure that is recorded in all cases is of the same type.

Each element, both in the train and test dataset, has 24 attributes. Three of the attributes are related to Operational Settings, and the rest are sensor measurements connected to the state of the machine. No additional information is provided in the description of the dataset to help describe the attributes.

The aim in the present research is to try to model the wear of the simulations of the test dataset to see how many work cycles remain until the error occurs.

\subsection{Data Description and Preprocessing}

The 3 Operational Settings attributes were not taken into account in the analysis. Thus, this study uses datasets containing 21 attributes that serve to describe the operation and behaviour of each simulation.

In addition, the simulations contained sensors that did not vary in time, and thus did not provide sensitive information about the behaviour. For that reason, these attributes were eliminated, and only those that showed changes over time were left. In that way, the number of useful variables was reduced from 24 to 14.

\section{Methodology and Data Preparation for Experiments}

Within Predictive Maintenance there are different components to be analysed. This study aims to demonstrate the usefulness of the AOI algorithm to detect changes (change detection), and later anomalies (AD), as well as to calculate the estimated time until the next behaviour error (RUL). The first step is to detect a change point, a moment in the behaviour in which the working process starts going outside control limits, since the term anomaly implies a failure.

To that end, it is important to follow a working action schedule, divided in two parts: (i) Training, and (ii) Testing.

\subsection{Train Methodology}

\begin{figure}[h!]
  \centering
  \includegraphics[width=0.8\textwidth]{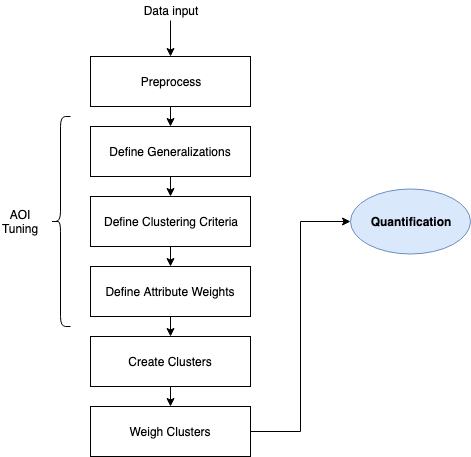}
  \caption{Part I of the Train Methodology of the Predictive Maintenance workflow with the AOI algorithm.}
  \label{train-methodology-part1}

\end{figure}

\begin{figure}[h!]
  \centering
  \includegraphics[width=0.8\textwidth]{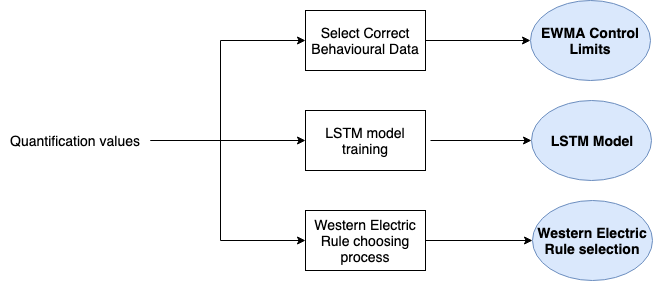}
  \caption{Part II of the Training Methodology of the Predictive Maintenance workflow with the AOI algorithm.}
  \label{train-methodology-part2}

\end{figure}

The main objective of the training phase is to construct the learning models which change detection and the RUL estimation are based on. The workflow is divided into two parts: (i) generation of the quantification function (Figure \ref{train-methodology-part1}), and (ii) construction of learning models (Figure \ref{train-methodology-part2}). 

One of the most important steps in any Data Analytics scenario is to inspect the data, as shown in the Data Description and Preprocessing section. On the one hand, the validity of the data must be ensured, since the results will greatly depend on that. On the other hand, this data must be understood, to be able to apply the most appropriate transformations according to its characteristics.

Next, the tuning process of the AOI algorithm begins. With the help of the knowledge of the domain experts, generalisations for the attributes must be established and the hierarchy-trees constructed. In the case of this experiment, there was no domain expert and the information offered by the open data source was limited. Thus, an alternative strategy to establish generalisations was followed. Since all the initial values of the dataset are numerical, the data was divided into blocks of 10 percentiles referring to each attribute. Thus, the arbitrariness of the division was reduced using a statistical approach.

Similarly, without domain experts that indicate the relevance of the attributes that characterise the data, the criterion for establishing the weights was arbitrary.

Once the AOI algorithm is tuned, the process of generating the clusters begins, assigning each one a weight according to the characteristics it has. The final set of generated clusters is denoted as the Knowledge Base. To calculate the weight of the clusters, it is first necessary to calculate the weight of the generalised Relationship. Depending on the level of generalisation of attributes in a generalised Relationship, the weight will vary. The more attributes that are generalised and the higher their level of generalisation, the lower the weight. This is because the quantification function indicates the degree of reliability which a state can be considered to be frequent in behaviour. If the level of generalisation is high, it indicates that the description of the state is more abstract, and therefore, the value of the cluster is lower. The partial weight of a cluster for a specific generalised Relationship is set out in Equation \ref{pesoparcial}.

\begin{equation}
\label{pesoparcial}
    generalised Level Weight (i) = \frac{\sum_{j=0}^{num attr}\frac{1}{2^{gen level}}}{num attr}
\end{equation}

The $numattr$ variable refers to the number of attributes of the dataset, and $genlevel$ indicates the generalisation level of that attribute. As explained in Figure \ref{concept-tree-aoi}, each attribute has its own hierarchy-tree, with distinct levels in which the values of the attribute are represented.

Table \ref{tableclusterpartial} shows an example of the values of the clusters after executing the AOI and assigning them the weight of the generalised Relationship. The first and the second columns refer to the id of the cluster and the weight of the generalised Relationship which the cluster belongs to. The third column indicates the number of elements that have made up the cluster. This variable is important when assigning the final weight to the clusters. The more instances that make up a group, the greater the weight in the quantification function. The fourth column indicates the number of elements existing before the cluster was formed. The last variable indicates the difference in weight that exists between the current generalised Relationship and the previous one.

\begin{table}[h]
\caption{Variables of interest to compute the cluster weights.}
\label{tableclusterpartial}
\begin{center}
\begin{tabular}{|c|c|c|c|c|}
\hline
Cluster	 & GenRel Weight & Elems & Outliers & Diff.\\
\hline
Cluster 1 & 0.8 & 100 & 500 & 0.2\\
\hline
Cluster 2 & 0.7 & 20 & 400 & 0.1\\
\hline
Cluster 3 & 0.55 & 10 & 380 & 0.15\\
\hline
... & ... & ... & ... & ...\\
\hline
Cluster N & 0.1 & 40 & 40 & 0.05\\
\hline
\end{tabular}
\end{center}
\end{table}

In Table \ref{tableclusterpartial} the partial weight of the clusters is shown taking into account only the generalised Relationship which they belong to. Equation \ref{pesototal} shows the method to calculate the final weight of the clusters. On the one hand, the weight of the generalised Relationship is computed, as shown in Equation \ref{pesoparcial}, and on the other it is necessary to calculate the value between that weight and the weight of the previous generalised Relationship that will be assigned. Hence, if the value of a generalised Relationship is 0.5, and the value of the preceding generalised Relationship is 0.6, the clusters of that generalised Relationship will have a value between 0.5 and 0.6 depending on how many elements it is composed of.

\begin{equation}
\label{pesototal}
    Cluster W (x, i) = GenLvW(i)  + \frac{inst(x))}{outl(i)}\cdot diffw(i)
\end{equation}

Each generalised Level can contain multiple clusters. Therefore in this equation, the weight for the $x$-th cluster on the $i$-th generalised Level is represented. $GenLvW(i)$ refers to the weight of the $i$-th generalised Level. $inst(x)$ indicates the instances that comprise the $x$-th cluster; $outl(i)$ makes reference to the number of outliers of the generalised Level $i$; and $diffw(i)$ is the difference of the weight between the $i$-th generalised Level and the previous one.

After establishing the weights for the clusters, the quantification function can be calculated. The quantification process consists of defining a numerical function that is correlated with the level of machine wear. In this case, the quantification was calculated for each of the simulations of the training dataset, as well as the test. To do this, the generalisation possibilities for each element of each simulation must first be analysed. Once the possible generalisations have been detected, a check is made to determine if there is a cluster that corresponds to that generalisation, and if it exists, the corresponding cluster weight is assigned. This process is explained further in the Test Methodology subsection.

Once the quantification is calculated, it is necessary to generate 3 more outputs to manage the change detection and RUL tasks: (i) EWMA control limits, (ii) LSTM predictive model, and (iii) the Western Electric Rule which best fits.

\subsubsection{Calculation of EWMA control limits}

To carry out the Change Detection, Statistical Process Control (SPC) charts were utilised. These charts are useful when there is an interest in detecting certain variability in a specific process. In this case, the so-called EWMA (Exponentially Weighted Moving Average) control chart was selected, which is particularly useful when small process shifts are of interest \cite{itsqc}.

\begin{figure}[h!]
  \centering
  \includegraphics[width=0.8\textwidth]{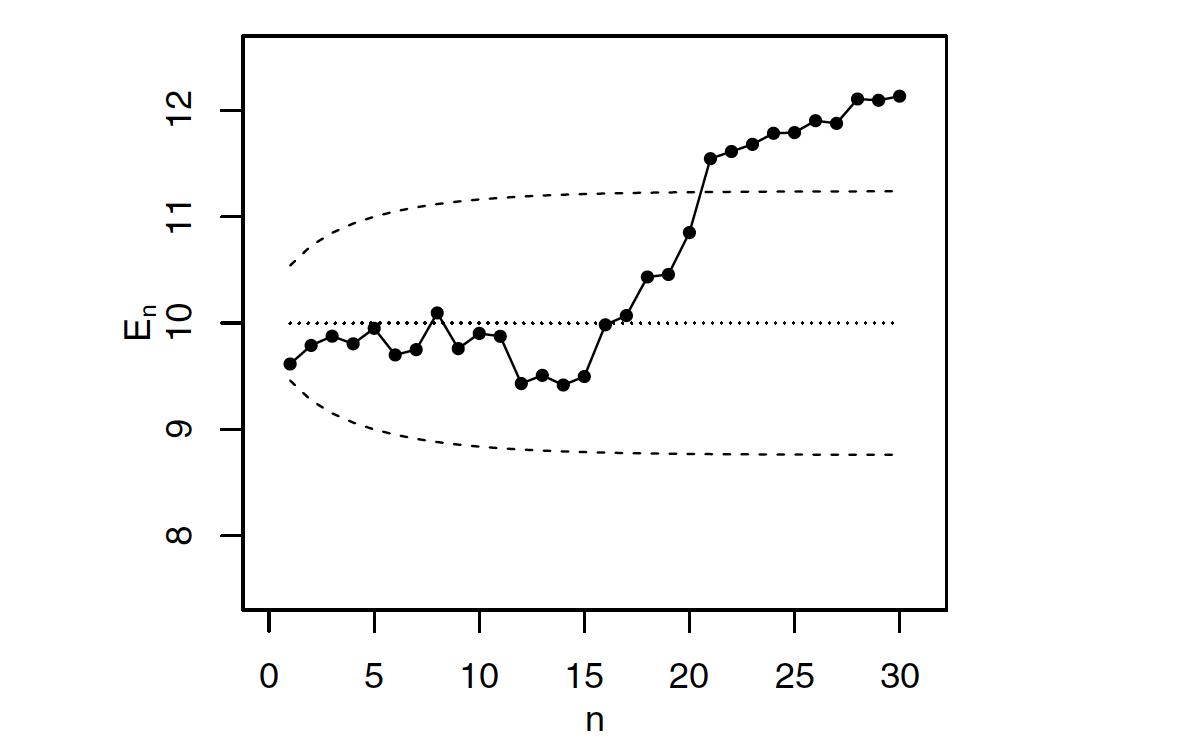}
  \caption{Example of an EWMA chart in which the UCL is exceeded \cite{itspc}.}
  \label{ewma-theoretical}

\end{figure}

The EWMA chart has two control limits, and once they are exceeded, it indicates that there has been a change in the behaviour of the system. In Figure \ref{ewma-theoretical} an example can be observed, and in Equations \ref{ucl} and \ref{lcl} the calculation of the control limits (Upper Control Limit: UCL, and Lower Control Limit: LCL) are shown. To generate the control thresholds of the EWMA chart, it is necessary to use data referring to the correct functioning of the machine. For this reason, a subset of initial instances of each simulation was selected.

\begin{equation}
\label{ucl}
    UCL = \mu_{0} + L\cdot \frac{\sigma }{\sqrt{n}}\cdot \sqrt{\frac{\lambda }{(2-\lambda)}\cdot (1-(1-\lambda)^{2i})}
\end{equation}

\begin{equation}
\label{lcl}
    LCL = \mu_{0} - L\cdot \frac{\sigma }{\sqrt{n}}\cdot \sqrt{\frac{\lambda }{(2-\lambda)}\cdot (1-(1-\lambda)^{2i})}
\end{equation}

$\mu_{0}$ and $\sigma$ refers to the mean and standard deviation of the population. $L$ is the multiple of the rational subgroup standard deviation that establishes the control limits. Its value is usually set between  2.7 and 3.0. $\lambda$ is the weight given to the most recent rational subgroup mean. It is recommended assigning a value between 0.05 and 0.25 \cite{itspc}.

\subsubsection{Generation of LSTM model}

A structure based on neural networks, specifically Long-Short Term Memory (LSTM), was used to predict the future states of a quantification function. The aim of this is to have a system that is capable of estimating the future behaviour of the quantification function. This leads to the calculation of the RUL.

\subsubsection{Selection of the Western Electric Rule that best fits}

Once a change in the behaviour of the process has been detected, it is necessary to predict how many work cycles remain until the error or breakdown occurs. Thus, it is necessary to establish a criterion to decide when a state is considered an anomaly.

In this study, the so-called Western Electric Rules (WER) were used. These are decision rules within the SPC area that are used to detect out-of-control conditions in control charts.

There are 4 WERs: 

(i) One point above Upper Control Limit or below Lower Control Limit. 

(ii) Two points above/below +2/-2 times standard deviation from the centre line. 

(iii) Three out of four points above/below +1/-1 time standard deviation from the centre line.

(iv) Eight points in a row above/below the centre line of the chart.

Depending on which WER is used, the condition to be able to detect an anomaly changes. 
In this study, the objective was to analyse the ability of each WER to detect an anomaly. It was considered that the point at which the error occurs is at the end of the simulation. A set of simulations in which the moment of failure is known was selected.  Then, the work cycle in which the failure occurred was compared with the estimation that each WER was able to provide for each simulation. To this end, the mean absolute error and the mean squared error were used as a criterion for measuring the quality of the prediction. The WER that best adjusted to the real moment of the failure was chosen. 

In this way, the training phase aims to generate 4 outputs to manage AD and RUL tasks. (i) Form the clusters and establish their weights to be able to represent the quantification function of each simulation. (ii) Calculate the control limits to represent the EWMA control chart which detects changes in the behaviour of the machine. (iii) Train an LSTM model capable of predicting the future behaviour of the quantification function, and (iv) select the WER that best suits the estimation of the remaining cycles until the failure happens.

\subsection{Test Methodology}

\begin{figure}[h!]
  \centering
  \includegraphics[width=0.8\textwidth, height=3.5cm]{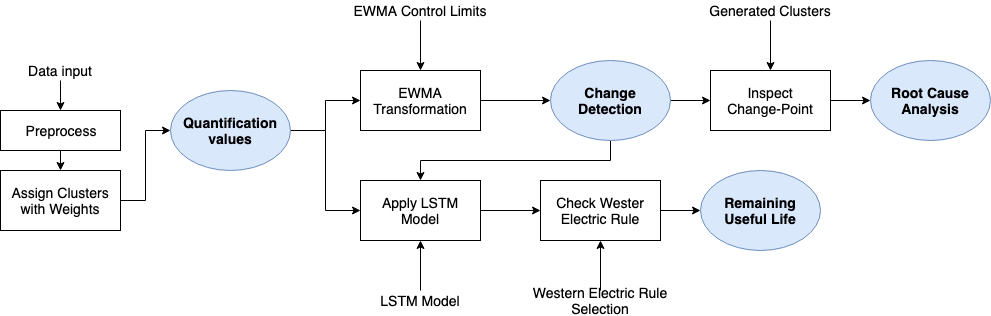}
  \caption{Test Methodology of the Predictive Maintenance workflow with the AOI algorithm.}
  \label{Methodology-test}

\end{figure}

In the test phase, the objective is to apply the models generated in the training phase, to assess their ability to detect changes in behaviour and predict the RUL. Figure \ref{Methodology-test} sets out the workflow.

The first step, after preprocessing the data, is to label the instances and assign them their corresponding weight. To do this, the attributes of each instance are inspected and whether there is a match in the Knowledge Base is determined. First, the clusters belonging to the generalised Relationship with the highest weight are inspected. In the case that there is no match for an element, its attributes are generalised to the next generalised Relationship with the greatest weight. At the end of this process, the instances of each simulation are associated with a Knowledge Base cluster and its respective weight. Hence, enabling the representation of quantification function of the current simulation. Instances that do not find equivalences in the Knowledge Base have a weight of zero, and are treated as anomalies.

Once the quantification function is represented for a simulation, the change detection and RUL processes can be executed.

\subsubsection{Change Detection}

To detect when a process is working outside the control limits, the EWMA control chart was used. Once the control thresholds are calculated in the training phase, it is necessary to transform the quantification function for the EWMA analysis. This transformation is performed by applying Equation \ref{ewma}.

\begin{equation}
    \label{ewma}
    z_{1} = \lambda \cdot x_{i} + (1-\lambda) \cdot z_{i-1}
\end{equation}

\begin{center}
    $z_{0} = \mu_{0}$
\end{center}

$x_{i}$ refers to the $i$-th element of the inspected dataset, and $z_{i-1}$ indicates the element of the EWMA transformation of the previous step.

It is therefore considered that a state is outside control limits when either of the two thresholds is exceeded.

\subsubsection{RUL}

Once the process is detected as working outside of control limits, the next step is to predict when the error may occur. To do so, it is important to take into account the following: the change detection process only finds states that work outside the control limits (change points), but this does not mean these states are always considered as anomalies.

The moment the system detects a change point, it is inspected to determine if the corresponding WER is fulfilled. In the case of compliance it is considered as an anomaly, and in the case of non compliance, the LSTM model is applied to predict the future behaviour of the quantification function until the condition indicated by the WER is met. Thus, the number of cycles that remain until the next anomaly occurrence is calculated.

\section{Results}

This experiment was performed with a minimum cluster size of 10, and decreasing iteratively as explained in the Variants of AOI section. In addition, 4 different concept-tree levels were defined for each attribute.

\begin{figure}[h!]
  \centering
  \includegraphics[width=0.8\textwidth]{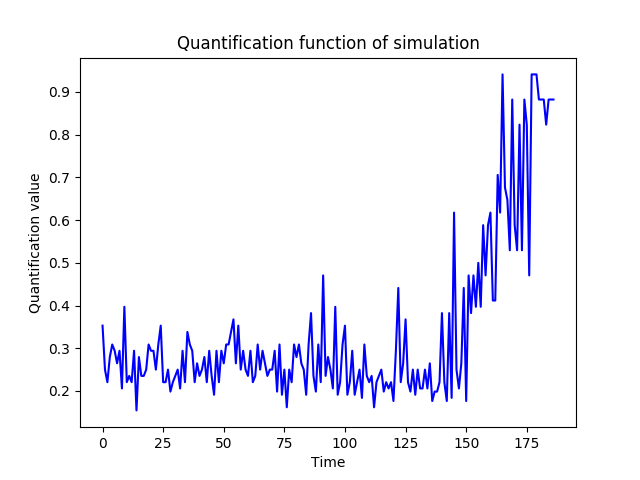}
  \caption{Example of a quantification function of a simulation.}
  \label{quantification-example}

\end{figure}

Figure \ref{quantification-example} shows an example in which the quantification function has been applied to a simulation. As can be seen, the initial values of the function are low, and at the end, these increase sharply. Considering that the type of failure for all simulations is the same in this dataset, we assume that the moment in which the values start to increase is when the faulty states become clear. The behaviour before the failure can be different for each simulation, but the states indicating the moment in which the failure starts are the same. For this reason the final part of the simulation has higher quantification values.

\begin{figure}[h!]
  \centering
  \includegraphics[width=0.8\textwidth]{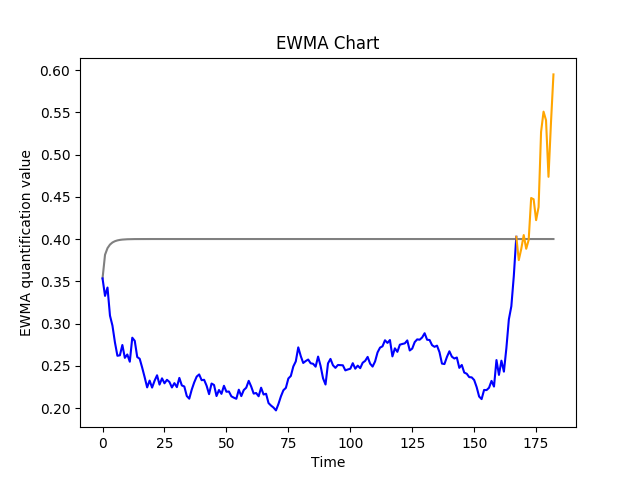}
  \caption{Representation of an EWMA chart for a simulation with the UCL.}
  \label{ewma-chart}

\end{figure}

The simulation was then transformed into the EWMA chart. Correct behavioural data was needed to compute the centre line and the control limits. Hence, assuming that the initial values of the quantification (i.e. until the function starts to increase) were not representative of failure states, these values were considered as correct. Thus, the first 100 points of each simulation were selected to calculate the control limits. Figure \ref{ewma-chart} shows the quantification function after being transformed into the EWMA chart.

As the moment of failure was considered to be when the quantification started to increase, the point on which the function exceeds the UCLwas inspected. Thus, only the UCL is represented in the EWMA chart. When the function achieved values higher than the UCL, change points that could lead to a failure were detected.

\begin{figure}[h!]
  \centering
  \includegraphics[width=0.8\textwidth]{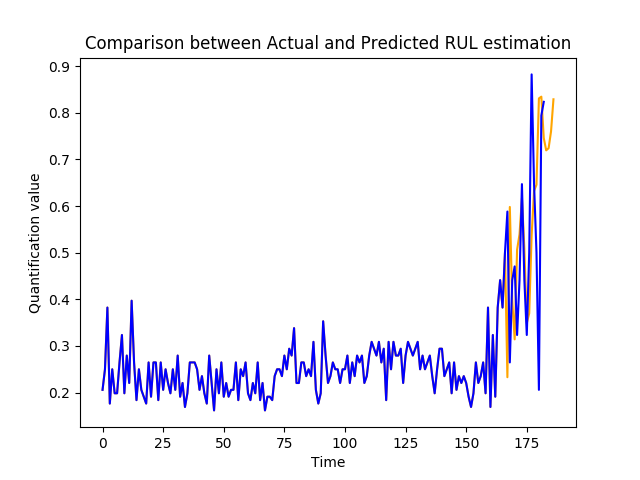}
  \caption{Comparison between Real and Predicted quantification function, starting from the change point.}
  \label{RUL-example}

\end{figure}

For the estimation of the RUL, an LSTM neural network was defined. To make the predictions, this LSTM model learnt from the values prior to when the first change point was detected. In Figure \ref{RUL-example} an example of the RUL prediction using the quantification function is shown. The blue section represents the real quantification function of a simulation, while the orange section is the prediction made by the LSTM model. The quality of the prediction model was measured by means of the root mean square error (RMSE) and the value was 0.036021.

Next, the WER that best fitted the real moment of failure was selected. The results of the tests performed for the different WERs are shown in Table \ref{tableresultswer}. As can be seen, WER 4 obtained the best result.

\begin{table}[h]
\caption{Representation of the mean error between the real RUL and the estimated RUL with the different WER conditions.}
\label{tableresultswer}
\begin{center}
\begin{tabular}{|c|c|c|}
\hline
WER & Mean Absolute Error & Mean Squared Error \\
\hline
WER 1 & 56.0 & 8518.48\\
\hline
WER 2 & 33.25 & 4125.35\\
\hline
WER 3 & 17.55 & 1084.13\\
\hline
WER 4 & 6.26 & 72.74\\
\hline

\end{tabular}
\end{center}
\end{table}

\begin{figure}[h!]
  \centering
  \includegraphics[width=0.8\textwidth]{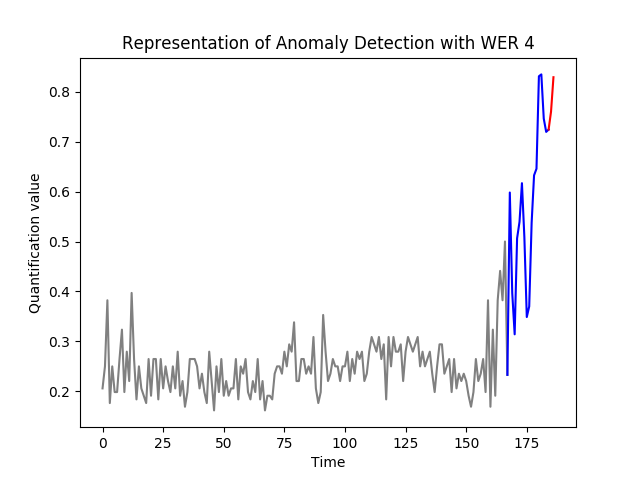}
  \caption{RUL prediction over a simulation using WER4 condition.}
  \label{wer4}

\end{figure}

Figure \ref{wer4} shows the example of a simulation starting from the detected change point, in which the future behaviour of the function was predicted and the fault was found. The grey section indicates the real quantification function until the change point was detected. The blue section represents the function predicted by means of the LSTM model, and the red section indicates the points of the function after the anomaly was found. The LSTM neural network was trained splitting the data into 70\% for the training phase and 30\% for the testing.

Finally, the quality of the implemented prediction system was measured. Two main aspects were taken into account:

(i) Mean absolute error: The difference between the real RUL value and the predicted value from the moment a change point was detected. The result of the analysis was 39.02.

(ii) Detection of the change point before the occurrence of the anomaly: Detecting if the simulation is going out of the control limits before the failure occurs is critical to prevent breakdowns. The change point was detected before the failure in 100\% of cases.

In some simulations the change point was detected very early (work cycle < 40). In these cases the prediction of the quantification function performed by the LSTM model was not as precise.  If only those simulations in which the change point occurs after the first 40 work cycles are taken into account, the result of the mean absolute error improves (13.30).

\section{Conclusions and Final Remarks}

This paper has presented a novel approach to a Predictive Maintenance scenario, using an Attribute Oriented Induction (AOI) algorithm. The results obtained confirm the validity of the proposal, as the methodology was able to detect anomalies, and calculate the RUL until breakage with a considerable degree of accuracy. As a result of this study, a number of conclusions can be drawn.

The study shows that it is possible to effectively implement a working methodology based on Data Driven Predictive Maintenance.  This is reflected in the choice of a method to detect anomalies, such as the WER, and its combination with neural network models to model the function and calculate the RUL. The error with which the WER is able to detect the breakage in the quantification function is very low, and thus ensures a more methodical process. Moreover, the modelling of algorithms over time, such as the LSTMs in this case, enables the detection of future anomalies to a high degree of precision.

One limitation that should be noted however, was that the use case presented in this paper did not include domain experts, despite this option being one of the clear advantages of the AOI algorithm.  Notwithstanding this limitation, the methodology proved valid to manage a Predictive Maintenance scenario, and the inclusion of domain expert knowledge in future studies would lead to improved results, since the experts were not assigned the weights and hierarchies of generalisation in the Knowledge Base. 

It is also important to consider that the value of the mean absolute error in the application of the WER is strictly related to the predictive capacity of the LSTM model. It is therefore expected that by improving the LSTM model, the predictive capacity of the system would be further improved. In this sense the LSTM model definition would benefit from more fine tuning, which in turn would improve the RUL prediction. 

Finally, it should be noted that, in future studies, the RCA calculation will be included in this methodology. The AOI algorithm has a strong descriptive power, which anticipates that the results can be considered very useful for domain experts.

\begin{acknowledgements}
This work has been developed by the intelligent systems for industrial systems group supported by the Department of Education, Language policy and Culture of the Basque Government. This work has been partially funded by the European Unions Horizon 2020 research and innovation programme project PROPHESY, under Grant Agreement no. 766994.
\end{acknowledgements}

%
%

\bibliographystyle{spbasic}      
\bibliography{template.bib}   

\end{document}